\documentclass{article}

\usepackage{ijcai16}
\usepackage{times}
\usepackage{url}
\usepackage{helvet}
\usepackage{courier}
\usepackage{amsfonts}
\usepackage{amsmath}
\usepackage{amssymb}
\usepackage{algorithm}
\usepackage{algorithmic}
\usepackage{multirow}
\usepackage{stmaryrd}
\usepackage{verbatim}
\usepackage[pdftex]{graphicx}
\usepackage{latexsym}
\usepackage{makecell}
\usepackage{array}

\title{
  Cost-aware Pre-training for Multiclass Cost-sensitive Deep Learning
}
\author{
  Yu-An Chung\\Department of CSIE\\National Taiwan University\\b01902040@ntu.edu.tw
  \And Hsuan-Tien Lin\\Department of CSIE\\National Taiwan University\\htlin@csie.ntu.edu.tw
  \And Shao-Wen Yang\\Intel Labs\\Intel Corporation\\shao-wen.yang@intel.com
}

\begin{document}

\maketitle

\begin{abstract}

Deep learning has been one of the most prominent machine learning techniques nowadays, being the state-of-the-art on a broad range of applications where automatic feature extraction is needed. Many such applications also demand varying costs for different types of mis-classification errors, but it is not clear whether or how such cost information can be incorporated into deep learning to improve performance. In this work, we first design a novel loss function that embeds the cost information for the training stage of cost-sensitive deep learning. We then show that the loss function can also be integrated into the pre-training stage to conduct cost-aware feature extraction more effectively. Extensive experimental results justify the validity of the novel loss function for making existing deep learning models cost-sensitive, and demonstrate that our proposed model with cost-aware pre-training and training outperforms non-deep models and other deep models that digest the cost information in other stages.

\end{abstract}

\section{Introduction}
\label{sec:introduction}

In many real-world machine learning applications \cite{tan1993cost,DBLP:conf/kdd/ChanS98,fan2000multiple,zhang2010cost,TJ2011}, classification errors may come with different costs; namely, some types of mis-classification errors may be (much) worse than others. For instance, when classifying bacteria \cite{TJ2011}, the cost of classifying a Gram-positive species as a Gram-negative one should be higher than the cost of classifying the species as another Gram-positive one because of the consequence on treatment effectiveness. Different costs are also useful for building a realistic face recognition system, where a government staff being mis-recognized as an impostor causes only little inconvenience, but an imposer mis-recognized as a staff can result in serious damage \cite{zhang2010cost}. It is thus important to take into account the de facto \emph{cost} of every type of error rather than only measuring the error rate and penalizing all types of errors equally.

The classification problem that mandates the learning algorithm to consider the cost information is called cost-sensitive classification. Amongst cost-sensitive classification algorithms, the binary classification ones \cite{elkan2001foundations,zadrozny2003cost} are somewhat mature with re-weighting the training examples \cite{zadrozny2003cost} being one major approach, while the multiclass classification ones are continuing to attract research attention \cite{domingos1999metacost,margineantu2001methods,abe2004iterative,tu2010one}. 

This work focuses on multiclass cost-sensitive classification, whose algorithms can be grouped into three categories \cite{abe2004iterative}. The first category makes the prediction procedure cost-sensitive \cite{kukar1998cost,domingos1999metacost,zadrozny2001learning}, generally done by equipping probabilistic classifiers with Bayes decision theory. The major drawback is that probability estimates can often be inaccurate, which in term makes cost-sensitive performance unsatisfactory. The second category makes the training procedure cost-sensitive, which is often done by transforming the training examples according to the cost information \cite{DBLP:conf/kdd/ChanS98,domingos1999metacost,zadrozny2003cost,beygelzimer2005error,langford2005sensitive}. However, the transformation step cannot take the particularities of the underlying classification model into account and thus sometimes has room for improvement. The third category specifically extends one particular classification model to be cost-sensitive, such as support vector machine \cite{tu2010one} or neural network \cite{kukar1998cost,zhou2006training}. Given that deep learning stands as an important class of models with its special properties to be discussed below, we aim to design cost-sensitive deep learning algorithms within the third category while borrowing ideas from other categories.

Deep learning models, or neural networks with deep architectures, are gaining increasing research attention in recent years. Training a deep neural network efficiently and effectively, however, comes with many challenges, and different models deal with the challenges differently. For instance, conventional fully-connected deep neural networks (DNN) generally initialize the network with an unsupervised pre-training stage before the actual training stage to avoid being trapped in a bad local minimal, and the unsupervised pre-training stage has been successfully carried out by stacked auto-encoders \cite{vincent2010stacked,dblp1829095,baldi2012autoencoders}. Deep belief networks \cite{hinton2006fast,le2008representational} shape the network as a generative model and commonly take restricted Boltzmann machines \cite{le2008representational} for pre-training. Convolutional neural networks (CNN) \cite{lecun1998gradient} mimic the visual perception process of human based on special network structures that result in less need for pre-training, and are considered the most effective deep learning models in tasks like image or speech recognition \cite{ciresan2011flexible,krizhevsky2012imagenet,abdel2014convolutional}.

While some existing works have studied cost-sensitive neural networks \cite{kukar1998cost,zhou2006training}, none of them have focused on cost-sensitive deep learning to the best of our knowledge. That is, we are the first to present cost-sensitive deep learning algorithms, with the hope of making deep learning more realistic for applications like bacteria classification and face recognition. In Section~\ref{sec:background}, we first formalize the cost-sensitive classification problem and review related deep learning works. Then, in Section~\ref{sec:csdl}, we start with a baseline algorithm that makes the prediction procedure cost-sensitive (first category). The features extracted from the training procedure of such an algorithm, however, are cost-blind. We then initiate a pioneering study on how the cost information can be digested in the training procedure (second category) of DNN and CNN. We design a novel loss function that matches the needs of neural network training while embedding the cost information. Furthermore, we argue that for DNN pre-trained with stacked auto-encoders, the cost information should not only be used for the training stage, but also the \textit{pre-training} stage. We then propose a novel pre-training approach for DNN (third category) that mixes unsupervised pre-training with a cost-aware loss function. Experimental results on deep learning benchmarks and standard cost-sensitive classification settings in Section~\ref{sec:experiment} verified that the proposed algorithm based on cost-sensitive training and cost-aware pre-training indeed yields the best performance, outperforming non-deep models as well as a broad spectrum of deep models that are either cost-insensitive or cost-sensitive in other stages. Finally, we conclude in Section~\ref{sec:conclusion}.

\section{Background}
\label{sec:background}

We will formalize the multiclass cost-sensitive classification problem before introducing deep learning and related works.

\subsection{Multiclass Cost-sensitive Classification}
We first introduce the multiclass classification problem and then extend it to the cost-sensitive setting. The $K$-class classification problem comes with a size-$N$ training set $S={\{(\mathbf{x}_{n}, y_{n})}\}^N_{n=1}$, where each input vector $\mathbf{x}_n$ is within an input space $\mathcal{X}$, and each label $y_{n}$ is within a label space $\mathcal{Y}=\{1,2,...,K\}$. The goal of multiclass classification is to train a classifier $g\colon \mathcal{X}\rightarrow\mathcal{Y}$ such that the expected error $\llbracket y\neq g(\mathbf{x})\rrbracket$ on test examples $(\mathbf{x}, y)$ is small.%
\footnote{$\llbracket\cdot \rrbracket$ is 1 when the inner condition is true, and 0 otherwise.}

Multiclass cost-sensitive classification extends multiclass classification by penalizing each type of mis-classification error differently based on some given costs. Specifically, consider a $K$ by $K$ cost matrix $\mathbf{C}$, where each entry $\mathbf{C}(y, k)\in [0, \infty)$ denotes the cost for predicting a class-$y$ example as class $k$ and naturally $\mathbf{C}(y, y) = 0$. The goal of cost-sensitive classification is to train a classifier~$g$ such that the expected cost $\mathbf{C}(y, g(\mathbf{x}))$ on test examples is small.

The cost-matrix setting is also called cost-sensitive classification with class-dependent costs. Another popular setting is to consider example-dependent costs, which means coupling an additional cost vector $\mathbf{c}\in[0, \infty)^K$ with each example $(\mathbf{x}, y)$, where the $k$-th component $\mathbf{c}[k]$ denotes the cost for classifying $\mathbf{x}$ as class $k$. During training, each $\mathbf{c}_{n}$ that accompanies $(\mathbf{x}_{n}, y_{n})$ is also fed to the learning algorithm to train a classifier $g$ such that the expected cost $\mathbf{c}[g(\mathbf{x})]$ is small with respect to the distribution that generates $(\mathbf{x}, y, \mathbf{c})$ tuples. The cost-matrix setting can be cast as a special case of the cost-vector setting by defining the cost vector in $(\mathbf{x}, y, \mathbf{c})$ as row $y$ of the cost matrix $\mathbf{C}$. In this work, we will eventually propose a cost-sensitive deep learning algorithm that works under the more general cost-vector setting.

\subsection{Neural Network and Deep Learning}
\label{sub:nndl}

There are many deep learning models that are successful for different applications nowadays \cite{lee2009convolutional,dblp1829095,ciresan2011flexible,krizhevsky2012imagenet,simonyan2014very}. In this work, we first study the fully-connected deep neural network (DNN) for multiclass classification as a starting point of making deep learning cost-sensitive. The DNN consists of $H$ hidden layer and parameterizes each layer $i\in \{1, 2, ..., H\}$ by $\theta_{i}=\{\mathbf{W}_{i}, \mathbf{b}_{i}\}$, where $\mathbf{W}_{i}$ is a fully-connected weight matrix and $\mathbf{b}_{i}$ is a bias vector that enter the neurons. That is, the weight matrix and bias vector applied on the input are stored within $\theta_{1}=\{\mathbf{W}_{1}, \mathbf{b}_{1}\}$. For an input feature vector~$\mathbf{ x}$, the~$H$ hidden layers of the DNN describe a complex feature transform function by computing $\phi(\mathbf{x})=s(\mathbf{W}_{H}\cdot s(\cdots s(\mathbf{W}_{2}\cdot s(\mathbf{W}_{1}\cdot \mathbf{x}+\mathbf{b}_{1})+\mathbf{b}_{2}) \cdots)+\mathbf{b}_{H})$, where $s(z)=\frac{1}{1+\exp(-z)}$ is the component-wise logistic function. Then, to perform multiclass classification, an extra softmax layer, parameterized by $\theta_{\mathrm{sm}}=\{\mathbf{W}_{\mathrm{sm}}, \mathbf{b}_{\mathrm{sm}}\}$, is placed after the $H$-th hidden layer. There are $K$ neurons in the softmax layer, where the $j$-th neuron comes with weights $\mathbf{W}_{\mathrm{sm}}^{(j)}$ and bias $\mathbf{b}_{\mathrm{sm}}^{(j)}$ and is responsible for estimating the probability of class $j$ given $\mathbf{x}$:
\begin{equation}
  \label{eq:softmax_output}
  P(y=j|\mathbf{x})=\frac{\exp(\phi(\mathbf{x})^{T}\mathbf{W}_{\mathrm{sm}}^{(j)}+\mathbf{b}_{\mathrm{sm}}^{(j)})}{\sum_{k=1}^{K}\exp(\phi(\mathbf{x})^{T}\mathbf{W}_{\mathrm{sm}}^{(k)}+\mathbf{b}_{\mathrm{sm}}^{(k)})}.
\end{equation}
Based on the probability estimates, the classifier trained from the DNN is naturally $g(\mathbf{x}) = \mbox{argmax}_{1 \le k \le K} P(y = k | \mathbf{x})$.

Traditionally, the parameters $\{\{\theta_{i}\}^{H}_{i=1}, \theta_{\mathrm{sm}}\}$ of the DNN are optimized by the back-propagation algorithm, which is essentially gradient descent, with respect to the negative log-likelihood loss function over the training set $S$:
\begin{equation}
  \label{eq:nll}
  L_{\mathrm{NLL}}(S)=\sum^N_{n=1}-\ln(P(y=y_{n}|\mathbf{x}_{n})).
\end{equation}

The strength of the DNN, through multiple layers of non-linear transforms, is to extract sophisticated features automatically and implement complex functions. However, the training of the DNN is non-trivial because of non-convex optimization and gradient diffusion problems, which degrade the test performance of the DNN when adding too many layers. \cite{hinton2006fast} first proposed a greedy layer-wise pre-training approach to solve the problem. The layer-wise pre-training approach performs a series of feature extraction steps from the bottom (input layer) to the top (last hidden layer) to capture higher level representations of original features along the network propagation.

In this work, we shall improve a classical yet effective unsupervised pre-training strategy, \emph{stacked denoising auto-encoders} \cite{vincent2010stacked}, for the DNN. Denoising auto-encoder (DAE) is an extension of regular auto-encoder. An auto-encoder is essentially a (shallow) neural network with one hidden layer, and consists of two parameter sets: $\{\mathbf{W}, \mathbf{b}\}$ for mapping the (normalized) input vector $\mathbf{x}\in [0, 1]^{d}$ to the $d'$-dimensional latent representation $\mathbf{h}$ by $\mathbf{h}=s(\mathbf{W}\cdot \mathbf{x}+\mathbf{b})\in [0, 1]^{d'}$; $\{\mathbf{W'}, \mathbf{b'}\}$ for reconstructing an input vector $\tilde{\mathbf{x}}$ from $\mathbf{h}$ by $\tilde{\mathbf{x}} = s(\mathbf{W'}\cdot \mathbf{h}+\mathbf{b'})$. The auto-encoder is trained by minimizing the total cross-entropy loss $L_{CE}(S)$ over $S$, defined as
\begin{equation}
 \label{eq:sum_cross_entropy}
 -\sum_{n=1}^N \sum_{j=1}^{d}\Bigl(\mathbf{x}_{n}[j]\ln \tilde{\mathbf{x}}_{n}[j]+(1-\mathbf{x}_{n}[j])\ln (1-\tilde{\mathbf{x}}_{n}[j])\Bigr),
\end{equation}
where $\mathbf{x}_{n}[j]$ denotes the $j$-th component of $\mathbf{x}_{n}$ and~$\tilde{\mathbf{x}}_{n}[j]$ is the corresponding reconstructed value.

The DAE extends the regular auto-encoder by randomly adding noise to inputs $\mathbf{x}_n$ before mapping to the latent representation, such as randomly setting some components of $\mathbf{x}_n$ to $0$. Several DAEs can then be stacked to form a deep network, where each layer receives its input from the latent representation of the previous layer. For the DNN, initializing with stacked DAEs is known to perform better than initializing with stacked regular auto-encoders \cite{vincent2010stacked} or initializing randomly. Below we will refer the DNN initialized with stacked DAEs and trained (fine-tuned) by back-propagation with~(\ref{eq:nll}) as the SDAE, while restricting the DNN to mean the model that is initialized randomly and trained with~(\ref{eq:nll}).

In this work, we will also extend another popular deep learning model, the convolutional neural network (CNN), for cost-sensitive classification. The CNN is based on a locally-connected network structure that mimics the visual perception process \cite{lecun1998gradient}. We will consider a standard CNN structure specified in Caffe\footnote{https://github.com/BVLC/caffe/tree/master/examples/cifar10} \cite{jia2014caffe}, which generally does not rely on a pre-training stage. Similar to the DNN, we consider the CNN with a softmax layer for multiclass classification.

\subsection{Cost-sensitive Neural Network}

Few existing works have studied cost-sensitive classification using neural networks \cite{kukar1998cost,zhou2006training}. \cite{zhou2006training} focused on studying the effect of sampling and threshold-moving to tackle the class imbalance problem using neural network as a core classifier rather than proposing general cost-sensitive neural network algorithms. \cite{kukar1998cost} proposed four approaches of modifying neural networks for cost-sensitivity. The first two approaches train a usual multiclass classification neural network, and then make the prediction stage of the trained network cost-sensitive by including the costs in the prediction formula; the third approach modifies the learning rate of the training algorithm base on the costs; the fourth approach, called MIN (minimization of the mis-classification costs), modifies the loss function of neural network training directly. Among the four proposed algorithms, MIN consistently achieves the lowest test cost \cite{kukar1998cost} and will be taken as one of our competitors. Nevertheless, none of the existing works, to the best of our knowledge, have conducted careful study on cost-sensitive algorithms for deep neural networks.

\section{Cost-sensitive Deep Learning}
\label{sec:csdl}

Before we start describing our proposed algorithm, we highlight a na{\"i}ve algorithm. For the DNN/SDAE/CNN that estimate the probability with (\ref{eq:softmax_output}), when given the full picture of the cost matrix, a cost-sensitive prediction can be obtained using Bayes optimal decision, which computes the expected cost of classifying an input vector $\mathbf{x}$ to each class and predicts the label that reaches the lowest expected cost:
\begin{equation}
  \label{eq:Bayes}
    g(\mathbf{x})=\underset{1\leqslant k\leqslant K}{\mathrm{argmin}}\sum_{y=1}^{K}P(y | \mathbf{x})\mathbf{C}(y, k).
\end{equation}
We will denote these algorithms as $\mathrm{DNN_{\mathrm{Bayes}}}$, $\mathrm{SDAE_{\mathrm{Bayes}}}$ and $\mathrm{CNN_{\mathrm{Bayes}}}$, respectively. These algorithms do not include the costs in the pre-training nor training stages. Also, those algorithms require knowing the full cost matrix, and cannot work under the cost-vector setting.

\subsection{Cost-sensitive Training}

The DNN essentially decomposes the multiclass classification problem to per-class probability estimation problems via the well-known
one-versus-all (OVA) decomposition. \cite{tu2010one} proposed the one-sided regression algorithm that extends OVA for support vector machine (SVM) to a cost-sensitive SVM by considering per-class regression problems. In particular, if regressors $r_k(\mathbf{x}) \approx \mathbf{c}[k]$ can be learned properly, a reasonable prediction can be made by
\begin{equation}
  \label{eq:pred_regression}
    g_{r}(\mathbf{x})\equiv\underset{1\leqslant k\leqslant K}{\mathrm{argmin}}\hspace{1mm}r_k(\mathbf{x}).
\end{equation}
\cite{tu2010one} further argued that the loss function of the regressor $r_k$ with respect to $\mathbf{c}[k]$ should be one-sided. That is,~$r_k(\mathbf{x})$ is allowed to underestimate the smallest cost $\mathbf{c}[y]$ and to overestimate other costs. Define $z_{n, k}=2\llbracket \mathbf{c}_{n}[k]=\mathbf{c}_{n}[y_{n}]\rrbracket -1$ for indicating whether~$\mathbf{c}_n[k]$ is the smallest within $\mathbf{c}_n$. The cost-sensitive SVM~\cite{tu2010one} minimizes a regularized version of the total one-sided loss $\xi_{n, k}=\mathrm{max}(z_{n, k}\cdot(r_{k}(\mathbf{x}_{n})-\mathbf{c}_{n}[k]), 0)$, where $r_k$ are formed by (kernelized) linear models. With such a design, the cost-sensitive SVM enjoys the following property \cite{tu2010one}:
\begin{equation} \label{eq:bounding_property}
\mathbf{c}_{n}[g_{r}(\mathbf{x}_{n})]\leqslant \sum_{k=1}^{K}\xi_{n, k}.
\end{equation}
That is, an upper bound $\sum_{k=1}^K \xi_{n, k}$ of the total cost paid by~$g_r$ on $\mathbf{x}_n$ is minimized within the cost-sensitive SVM.

If we replace the softmax layer of the DNN or the CNN with regression outputs (using the identity function instead of the logistic one for outputting), we can follow~\cite{tu2010one} to make DNN and CNN cost-sensitive by letting each output neuron estimate $\mathbf{c}[k]$ as $r_k$ and predicting with (\ref{eq:pred_regression}). The training of the cost-sensitive DNN and CNN can also be done by minimizing the total one-sided loss. Nevertheless, the one-sided loss is not differentiable at some points, and back-propagation (gradient descent) cannot be directly applied. We thus derive a \emph{smooth} approximation of $\xi_{n, k}$ instead. Note that the new loss function should not only approximate~$\xi_{n, k}$ but also be an upper bound of $\xi_{n, k}$ to keep enjoying the bounding property of (\ref{eq:bounding_property}). \cite{lm:99} has shown a smooth approximation $u + \frac{1}{ \alpha} \cdot \ln (1 + \exp(- \alpha u)) \approx \max(u, 0)$ when deriving the smooth SVM. Taking $\alpha=1$ leads to $\mathrm{LHS} = \ln(1 + \exp(u))$, which is trivially an upper bound of $\max(u, 0)$ because $\ln(1+\exp(u)) > u$, and $\ln(1+\exp(u)) > \ln(1) = 0$. Based on the approximation, we define
\begin{equation}
  \label{eq:smooth_loss}
  \delta_{n, k} \equiv \ln(1+\exp(z_{n, k}\cdot (r_{k}(\mathbf{x}_{n})-\mathbf{c}_{n}[k]))).
\end{equation}
$\delta_{n, k}$ is not only a smooth approximation of $\xi_{n, k}$ that enjoys the differentiable property, but also an upper bound of $\xi_{n, k}$ to keep the bounding property of (\ref{eq:bounding_property}) held. That is, we can still ensure a small total cost by minimizing the newly defined smooth one-sided regression ($\mathrm{SOSR}$) loss over all examples:
\begin{equation}
  \label{eq:sum_smooth_loss}
    L_{\mathrm{SOSR}}(S)=\sum_{n=1}^{N}\sum_{k=1}^{K}\delta_{n, k}.
\end{equation}

We will refer to these algorithms, which replace the softmax layer of the DNN/SDAE/CNN with a regression layer parameterized by $\theta_{\mathrm{SOSR}}=\{\mathbf{W}_{\mathrm{SOSR}}, \mathbf{b}_{\mathrm{SOSR}}\}$ and minimize~(\ref{eq:sum_smooth_loss}) with back-propagation, as $\mathrm{DNN}_{\mathrm{SOSR}}$, $\mathrm{SDAE}_{\mathrm{SOSR}}$ and $\mathrm{CNN}_{\mathrm{SOSR}}$. These algorithms work with the cost-vector setting. They include costs in the training stage, but not the pre-training stage.

\subsection{Cost-aware Pre-training}

For multiclass classification, the pre-training stage, either in a totally unsupervised or partially supervised manner \cite{Bengio-nips-2006}, has been shown to improve the performance of the DNN and several other deep models \cite{Bengio-nips-2006,hinton2006fast,erhan2010does}. The reason is that pre-training usually helps initialize a neural network with better weights that prevent the network from getting stuck in poor local minima. In this section, we propose a cost-aware pre-training approach that leads to a novel cost-sensitive deep neural network (CSDNN) algorithm.

CSDNN is designed as an extension of $\mathrm{SDAE}_{\mathrm{SOSR}}$. Instead of pre-training with SDAE, CSDNN takes stacked \emph{cost-sensitive auto-encoders} (CAE) for pre-training instead. For a given cost-sensitive example $(\mathbf{x}, y, \mathbf{c})$, CAE tries not only to denoise and reconstruct the original input $\mathbf{x}$ like DAE, but also to digest the cost information by \emph{reconstructing the cost vector} $\mathbf{c}$. That is, in addition to $\{\mathbf{W}, \mathbf{b}\}$ and $\{\mathbf{W'}, \mathbf{b'}\}$ for DAE, CAE introduces an extra parameter set $\{\mathbf{W''}, \mathbf{b''}\}$ fed to regression neurons from the hidden representation. Then, we can mix the two loss functions $L_{\mathrm{CE}}$ and $L_{\mathrm{SOSR}}$ with a balancing coefficient $\beta\in [0, 1]$, yielding the following loss function for CAE over $S$:
\begin{equation}
  \label{eq:cae_loss}
    L_{\mathrm{CAE}}(S) = (1-\beta) \cdot L_{\mathrm{CE}}(S) + \beta \cdot L_{\mathrm{SOSR}}(S)
\end{equation}
The mixture step is a widely-used technique for multi-criteria optimization \cite{moobook}, where $\beta$ controls the balance between reconstructing the original input $\mathbf{x}$ and the cost vector $\mathbf{c}$. A positive $\beta$ makes CAE cost-aware during its feature extraction, while a zero $\beta$ makes CAE degenerate to DAE. Similar to DAEs, CAEs can then be stacked to initialize a deep neural network before the weights are fine-tuned by back-propagation with (\ref{eq:sum_smooth_loss}). The resulting algorithm is named CSDNN, which is cost-sensitive in both the pre-training stage (by CAE) and the training stage (by (\ref{eq:sum_smooth_loss})), and can work under the general cost-vector setting. The full algorithm is listed in Algorithm~\ref{alg:CSDNN}.
\begin{algorithm}[h]
  \renewcommand{\algorithmicrequire}{\textbf{Input:}}
  \renewcommand{\algorithmicensure}{\textbf{Output:}}
  \caption{CSDNN}
  \label{alg:CSDNN}
  \begin{algorithmic}[1]
  \REQUIRE Cost-sensitive training set $S={\{(\mathbf{x}_{n}, y_{n}, \mathbf{c}_{n})}\}^N_{n=1}$
  \FOR{each hidden layer $\theta_{i}=\{\mathbf{W}_{i}, \mathbf{b}_{i}\}$}
      \STATE Learn a CAE by minimizing (\ref{eq:cae_loss}).
      \STATE Take $\{\mathbf{W}_{i}, \mathbf{b}_{i}\}$ of CAE as $\theta_{i}$.
  \ENDFOR
  \STATE Fine-tune the network parameters $\{\{\theta_{i}\}^{H}_{i=1}, \theta_{\mathrm{SOSR}}\}$ by minimizing (\ref{eq:sum_smooth_loss}) using back-propagation.
  \ENSURE The fine-tuned deep neural network with (\ref{eq:pred_regression}) as $g_{r}$.
  \end{algorithmic}
\end{algorithm}

CSDNN is essentially $\mathrm{SDAE}_{\mathrm{SOSR}}$ with DAEs replaced by CAEs with the hope of more effective cost-aware feature extraction. We can also consider $\mathrm{SCAE}_{\mathrm{Bayes}}$ which does the same for $\mathrm{SDAE}_{\mathrm{Bayes}}$. The CNN, due to its special network structure, generally does not rely on stacked DAEs for pre-training, and hence cannot be extended by stacked CAEs.

As discussed, DAE is a degenerate case of CAE. Another possible degeneration is to consider CAE with less complete cost information. For instance, a na{\"i}ve cost vector defined by $\mathbf{\hat{c}}_n[k] = \llbracket y_n \neq k \rrbracket$ encodes the label information (whether the prediction is erroneous with respect to the demanded label) but not the complete cost information. To study whether it is necessary to take the complete cost information into account in CAE, we design two variant algorithms that replace the cost vectors in CAEs with $\mathbf{\hat{c}}_n[k]$, which effectively makes those CAEs \emph{error-aware}. Then, $\mathrm{SCAE}_{\mathrm{Bayes}}$ becomes $\mathrm{SEAE}_{\mathrm{Bayes}}$ (with E standing for error); CSDNN becomes $\mathrm{SEAE}_{\mathrm{SOSR}}$.

\section{Experiments}
\label{sec:experiment}

In the previous section, we have derived many cost-sensitive deep learning algorithms, each with its own specialty. They can be grouped into two series: those minimizing with~(\ref{eq:nll}) and predicting with~(\ref{eq:Bayes}) are $\mathrm{Bayes}$-series algorithms ($\mathrm{DNN}_{\mathrm{Bayes}}$, $\mathrm{SDAE}_{\mathrm{Bayes}}$, $\mathrm{SEAE}_{\mathrm{Bayes}}$, $\mathrm{SCAE}_{\mathrm{Bayes}}$, and $\mathrm{CNN}_{\mathrm{Bayes}}$); those minimizing with~(\ref{eq:sum_smooth_loss}) and predicting with~(\ref{eq:pred_regression}) are $\mathrm{SOSR}$-series algorithms ($\mathrm{DNN}_{\mathrm{SOSR}}$, $\mathrm{SDAE}_{\mathrm{SOSR}}$, $\mathrm{SEAE}_{\mathrm{SOSR}}$, $\mathrm{CSDNN} \equiv \mathrm{SCAE}_{\mathrm{SOSR}}$, $\mathrm{CNN}_{\mathrm{SOSR}}$). Note that the $\mathrm{Bayes}$-series can only be applied to the cost-matrix setting while the $\mathrm{SOSR}$-series can deal with the cost-vector setting. The two series help understand whether it is beneficial to consider the cost information in the training stage.

Within each series, $\mathrm{CNN}$ is based on a locally-connected structure, while $\mathrm{DNN}$, $\mathrm{SDAE}$, $\mathrm{SEAE}$ and $\mathrm{SCAE}$ are fully-connected and differ by how pre-training is conducted, ranging from none, unsupervised, error-aware, to cost-aware. The wide range helps understand the effectiveness of digesting the cost information in the pre-training stage.

Next, the two series will be compared with the $\mathrm{blind}$-series algorithms ($\mathrm{DNN}_{\mathrm{blind}}$, $\mathrm{SDAE}_{\mathrm{blind}}$, and $\mathrm{CNN}_{\mathrm{blind}}$), which are the existing algorithms that do not incorporate the cost information at all, to understand the importance of taking the cost information into account. The two series will also be compared against two baseline algorithms: $\mathrm{CSOSR}$ \cite{tu2010one}, a non-deep algorithm that our proposed $\mathrm{SOSR}$-series originates from; $\mathrm{MIN}$ \cite{kukar1998cost}, a neural-network algorithm that is cost-sensitive in the training stage like the $\mathrm{SOSR}$-series but with a different loss function. The algorithms along with highlights on where the cost information is digested are summarized in Table~\ref{tab:models_comp}.

\makeatletter
\newcommand{\thickhline}{
    \noalign {\ifnum 0=`}\fi \hrule height 1pt
    \futurelet \reserved@a \@xhline
}
\newcolumntype{"}{@{\hskip\tabcolsep\vrule width 1.5pt\hskip\tabcolsep}}
\makeatother

\begin{table*}[t]
  \caption{cost-awareness of algorithms (O: cost-aware; E: error-aware; X: cost-blind)}
  \label{tab:models_comp}
  \centering
  \small
  \resizebox{\textwidth}{!}{
    \begin{tabular}{|c"c|c|c||c|c||c|c|c|c|c||c|c|c|c|c|} \hline
    \diaghead{\theadfont Row ColumnRow}{Stage}{Algorithm}                      
    &  $\mathrm{DNN}_{\mathrm{blind}}$  &  $\mathrm{SDAE}_{\mathrm{blind}}$  &  $\mathrm{CNN}_{\mathrm{blind}}$  &  $\mathrm{CSOSR}$                 &  $\mathrm{MIN}$
    &  $\mathrm{DNN}_{\mathrm{Bayes}}$  &  $\mathrm{SDAE}_{\mathrm{Bayes}}$  &  $\mathrm{SEAE}_{\mathrm{Bayes}}$ &  $\mathrm{SCAE}_{\mathrm{Bayes}}$ &  $\mathrm{CNN}_{\mathrm{Bayes}}$
    &  $\mathrm{DNN}_{\mathrm{SOSR}}$   &  $\mathrm{SDAE}_{\mathrm{SOSR}}$   &  $\mathrm{SEAE}_{\mathrm{SOSR}}$  &  $\mathrm{CSDNN}$                 &  $\mathrm{CNN}_{\mathrm{SOSR}}$  \\\thickhline
    pre-training    &    none    &    X    &    none  &    none    &    none    &    none    &    X    &    E    &    O    &    none    &    none    &    X    &    E    &    O    &    none  \\\hline
    training        &     X      &    X    &     X    &     O      &     O      &     X      &    X    &    X    &    X    &     X      &     O      &    O    &    O    &    O    &     O    \\\hline
    prediction      &     X      &    X    &     X    &     X      &     X      &     O      &    O    &    O    &    O    &     O      &     X      &    X    &    X    &    X    &     X    \\\hline
    \end{tabular}
  }
\end{table*}

\subsection{Setup}

We conducted experiments on MNIST, bg-img-rot (the hardest variant of MNIST provided in \cite{larochelle2007empirical}), SVHN \cite{netzer2011reading}, and CIFAR-10 \cite{Krizhevsky09}. The first three datasets belong to handwritten digit recognition and aim to classify each image into a digit of $0$ to $9$ correctly; CIFAR-10 is a well-known image recognition dataset which contains 10 classes such as car, ship and animal. For all four datasets, the training, validation, and testing split follows the source websites; the input vectors in the training set are linearly scaled to $[0, 1]$, and the input vectors in the validation and testing sets are scaled accordingly.

The four datasets are originally collected for multiclass classification and contain no cost information. We adopt the most frequently-used benchmark in cost-sensitive learning, the randomized proportional setup \cite{abe2004iterative}, to generate the costs. The setup is for the cost-matrix setting. It first generates a $K \times K$ matrix $\mathbf{C}$, and sets the diagonal entries $\mathbf{C}(y, y)$ to $0$ while sampling the non-diagonal entries $\mathbf{C}(y, k)$ uniformly from $[0, 10\frac{|\{n:y_{n}=k\}|}{|\{n:y_{n}=y\}|}]$. The randomized proportional setup generates the cost information that takes the class distribution of the dataset into account, charging a higher cost (in expectation) for mis-classifying a minority class, and can thus be used to deal with imbalanced classification problems. Note that we take this benchmark cost-matrix setting to give prediction-stage cost-sensitive algorithms like the $\mathrm{Bayes}$-series a fair chance of comparison. We find that the range of the costs can affect the numerical stability of the algorithms, and hence scale all the costs by the maximum value within $\mathbf{C}$ during training in our implementation. The reported test results are based on the unscaled $\mathbf{C}$.

Arguably one of the most important use of cost-sensitive classification is to deal with imbalanced datasets. Nevertheless, the four datasets above are somewhat balanced, and 
the randomized proportional setup may generate similar cost for each type of mis-classification error. To better meet the real-world usage scenario, we further conducted experiments to evaluate the algorithms with imbalanced datasets. In particular, for each dataset, we construct a variant dataset by randomly picking four classes and removing $70\%$ of the examples that belong to those four classes. We will name these imbalanced variants as $\mathrm{MNIST}_{\mathrm{imb}}$, $\mathrm{bg}$-$\mathrm{img}$-$\mathrm{rot}_{\mathrm{imb}}$, $\mathrm{SVHN}_{\mathrm{imb}}$, and $\mathrm{CIFAR}$-$\mathrm{10}_{\mathrm{imb}}$, respectively.

All experiments were conducted using Theano. For algorithms related to DNN and SDAE, we selected the hyperparameters by following \cite{vincent2010stacked}. The $\beta$ in (\ref{eq:cae_loss}), needed by $\mathrm{SEAE}$ and $\mathrm{SCAE}$ algorithms, was selected among $\{0, 0.05, 0.1, 0.25, 0.4, 0.75, 1\}$. As mentioned, 
for CNN, we considered a standard structure in Caffe \cite{jia2014caffe}.

\subsection{Experimental Results}
\label{sec:experimental results}

The average test cost of each algorithm along with the standard error is shown in Table~\ref{tab:result}. The best result\footnote{The selected $\mathrm{CSDNN}$ that achieved the test cost listed in Table~\ref{tab:result} is composed of 3 hidden layers, and each hidden layer consists of 3000 neurons.} per dataset among all algorithms is highlighted in bold.

\begin{table*}[t]
  \caption{Average test cost}
  \label{tab:result}
  \centering
  \small
  \resizebox{\textwidth}{!}{
    \begin{tabular}{|c"c|c|c||c|c||c|c|c|c|c||c|c|c|c|c|} \hline
    \diaghead{\theadfont Diag ColumnRow}{Dataset}{Algorithm}
    &  $\mathrm{DNN}_{\mathrm{blind}}$  &  $\mathrm{SDAE}_{\mathrm{blind}}$  &  $\mathrm{CNN}_{\mathrm{blind}}$  &  $\mathrm{CSOSR}$                  &  $\mathrm{MIN}$
    &  $\mathrm{DNN}_{\mathrm{Bayes}}$  &  $\mathrm{SDAE}_{\mathrm{Bayes}}$  &  $\mathrm{SEAE}_{\mathrm{Bayes}}$ &  $\mathrm{SCAE}_{\mathrm{Bayes}}$  &  $\mathrm{CNN}_{\mathrm{Bayes}}$
    &  $\mathrm{DNN}_{\mathrm{SOSR}}$   &  $\mathrm{SDAE}_{\mathrm{SOSR}}$   &  $\mathrm{SEAE}_{\mathrm{SOSR}}$  &  $\mathrm{CSDNN}$                  &  $\mathrm{CNN}_{\mathrm{SOSR}}$  \\\thickhline
    MNIST
            &  $0.11\pm 0.00$  &  $0.10\pm 0.00$           &  $0.10\pm 0.00$   &  $0.10\pm 0.00$           &  $0.10\pm 0.003$
            &  $0.10\pm 0.00$  &  $\mathbf{0.09\pm 0.00}$  &  $0.09\pm 0.00$  &  $0.09\pm 0.00$           &  $\mathbf{0.09\pm 0.00}$
            &  $0.10\pm 0.00$  & $0.09\pm 0.00$            &  $0.09\pm 0.00$  &  $\mathbf{0.09\pm 0.00}$  &  $\mathbf{0.08\pm 0.00}$  \\\hline
    bg-img-rot 
            &  $3.33\pm 0.06$  &  $3.28\pm 0.07$  &  $3.05\pm 0.07$  &  $3.25\pm 0.06$           &  $3.02\pm 0.06$
            &  $2.95\pm 0.07$  &  $2.66\pm 0.07$  &  $2.85\pm 0.07$  &  $2.54\pm 0.07$           &  $2.40\pm 0.07$
            &  $3.21\pm 0.07$  &  $2.99\pm 0.07$  &  $3.00\pm 0.07$  &  $\mathbf{2.34\pm 0.07}$  & $\mathbf{2.29\pm 0.07}$            \\\hline
    SVHN
            &  $1.58\pm 0.03$  &  $1.40\pm 0.03$  &  $0.91\pm 0.03$  &  $1.17\pm 0.03$           &  $1.19\pm 0.03$
            &  $1.07\pm 0.03$  &  $0.93\pm 0.03$  &  $0.94\pm 0.03$  &  $0.88\pm 0.03$           &  $\mathbf{0.85\pm 0.03}$ 
            &  $1.02\pm 0.03$  &  $0.92\pm 0.03$  &  $0.99\pm 0.03$  &  $\mathbf{0.83\pm 0.03}$  &  $\mathbf{0.82\pm 0.03}$           \\\hline
    CIFAR-10
            &  $3.46\pm 0.04$  &  $3.26\pm 0.05$  &  $2.51\pm 0.04$  &  $3.30\pm 0.04$           &  $3.19\pm 0.05$
            &  $2.80\pm 0.05$  &  $2.52\pm 0.05$  &  $2.68\pm 0.05$  &  $2.38\pm 0.04$           &  $2.34\pm 0.05$
            &  $2.74\pm 0.05$  &  $2.48\pm 0.04$  &  $2.52\pm 0.05$  &  $\mathbf{2.24\pm 0.05}$  &  $\mathbf{2.25\pm 0.04}$           \\\hline
    $\mathrm{MNIST}_{\mathrm{imb}}$ 
            &  $0.32\pm 0.01$  &  $0.31\pm 0.01$  &  $0.19\pm 0.01$  &  $0.26\pm 0.01$           &  $0.27\pm 0.01$ 
            &  $0.23\pm 0.01$  &  $0.20\pm 0.01$  &  $0.20\pm 0.01$  &  $\mathbf{0.18\pm 0.01}$  &  $\mathbf{0.18\pm 0.01}$
            &  $0.22\pm 0.01$  &  $0.20\pm 0.01$  &  $0.19\pm 0.01$  &  $\mathbf{0.17\pm 0.01}$  &  $\mathbf{0.17\pm 0.01}$           \\\hline
    $\mathrm{bg}$-$\mathrm{img}$-$\mathrm{rot}_{\mathrm{imb}}$
            &  $15.9\pm 0.70$  &  $13.8\pm 0.70$  &  $5.04\pm 0.67$  &  $8.55\pm 0.70$           &  $8.40\pm 0.69$
            &  $7.19\pm 0.69$  &  $5.10\pm 0.70$  &  $4.95\pm 0.70$  &  $4.73\pm 0.70$           &  $\mathbf{4.49\pm 0.68}$
            &  $6.89\pm 0.70$  &  $4.99\pm 0.69$  &  $4.86\pm 0.69$  &  $\mathbf{4.16\pm 0.68}$  &  $\mathbf{4.39\pm 0.69}$           \\\hline
    $\mathrm{SVHN}_{\mathrm{imb}}$
            &  $1.79\pm 0.01$  &  $1.60\pm 0.01$  &  $0.31\pm 0.01$  &  $1.05\pm 0.01$           &  $0.99\pm 0.01$
            &  $0.53\pm 0.01$  &  $0.33\pm 0.01$  &  $0.34\pm 0.01$  &  $0.29\pm 0.01$           &  $0.28\pm 0.01$
            &  $0.51\pm 0.01$  &  $0.31\pm 0.01$  &  $0.31\pm 0.01$  &  $\mathbf{0.26\pm 0.01}$  &  $0.28\pm 0.01$                    \\\hline
    $\mathrm{CIFAR}$-$\mathrm{10}_{\mathrm{imb}}$
            &  $19.1\pm 0.09$  &  $17.7\pm 0.09$  &  $7.29\pm 0.08$  &  $10.1\pm 0.09$  &           $11.2\pm 0.09$
            &  $8.16\pm 0.09$  &  $7.48\pm 0.09$  &  $7.25\pm 0.08$  &  $6.97\pm 0.09$  &           $6.81\pm 0.09$
            &  $7.86\pm 0.08$  &  $7.44\pm 0.09$  &  $7.14\pm 0.09$  &  $\mathbf{6.48\pm 0.09}$  &  $6.63\pm 0.08$                    \\\hline
    \end{tabular}
  }
\end{table*}

\newcommand{\myparagraph}[1]{\vspace{.4em} \noindent \textbf{#1}\ }
\myparagraph{Is it necessary to consider costs?}
$\mathrm{DNN}_{\mathrm{blind}}$ and $\mathrm{SDAE}_{\mathrm{blind}}$ performed the worst on almost all the datasets. While $\mathrm{CNN}_{\mathrm{blind}}$ was slightly better than those two, it never reached the best performance for any dataset. The results indicate the necessity of taking the cost information into account.

\myparagraph{Is it necessary to go deep?}
The two existing cost-sensitive baselines, $\mathrm{CSOSR}$ and $\mathrm{MIN}$, outperformed the cost-blind algorithms often, but were usually not competitive to cost-sensitive deep learning algorithms. The results validate
the importance of studying cost-sensitive deep learning.

\myparagraph{Is it necessary to incorporate costs during training?}
$\mathrm{SOSR}$-series models, especially under the imbalanced scenario, generally outperformed their $\mathrm{Bayes}$ counterparts. The results demonstrate the usefulness of the proposed (\ref{eq:smooth_loss}) and (\ref{eq:sum_smooth_loss}) and the importance of incorporating the cost information during the training stage.

\myparagraph{Is it necessary to incorporate costs during pre-training?}
$\mathrm{CSDNN}$ outperformed both $\mathrm{SEAE}_{\mathrm{SOSR}}$ and $\mathrm{SDAE}_{\mathrm{SOSR}}$, and $\mathrm{SDAE}_{\mathrm{SOSR}}$ further outperformed $\mathrm{DNN}_{\mathrm{SOSR}}$. The results show that for the fully-connected structure where pre-training is needed, our newly proposed cost-aware pre-training with CAE is indeed helpful in making deep learning cost-sensitive.

\myparagraph{Which is better, $\mathrm{CNN}_{\mathrm{SOSR}}$ or $\mathrm{CSDNN}$?}
The last two columns in Table~\ref{tab:result} show that $\mathrm{CSDNN}$ is competitive to $\mathrm{CNN}_{\mathrm{SOSR}}$, with both algorithms usually achieving  the best performance. $\mathrm{CSDNN}$ is slightly better on two datasets. Note that CNNs are known to be powerful for image recognition tasks, which match the datasets that we have used. Hence, it is not surprising that CNN can reach promising performance with our proposed $\mathrm{SOSR}$ loss (\ref{eq:sum_smooth_loss}). Our efforts not only make CNN cost-sensitive, but also result in the CSDNN algorithm that makes the full-connected deep neural network cost-sensitive with the help of cost-aware pre-training via $\mathrm{CAE}$.

\begin{figure}[t]
\includegraphics[scale=0.36]{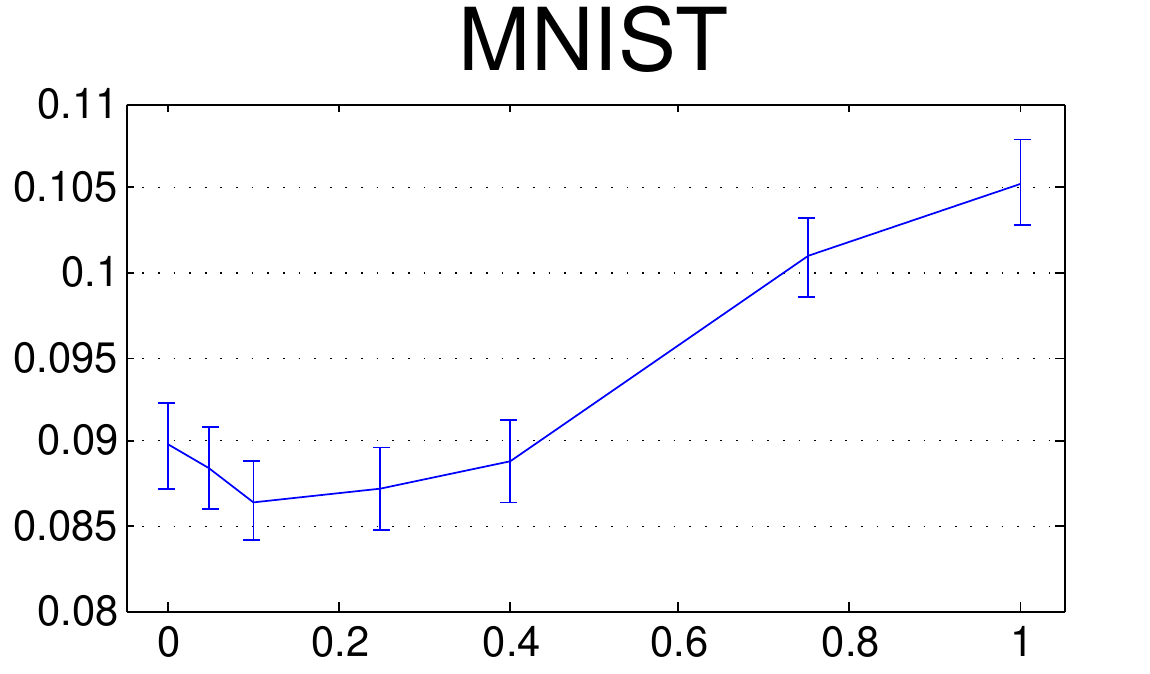}
\includegraphics[scale=0.36]{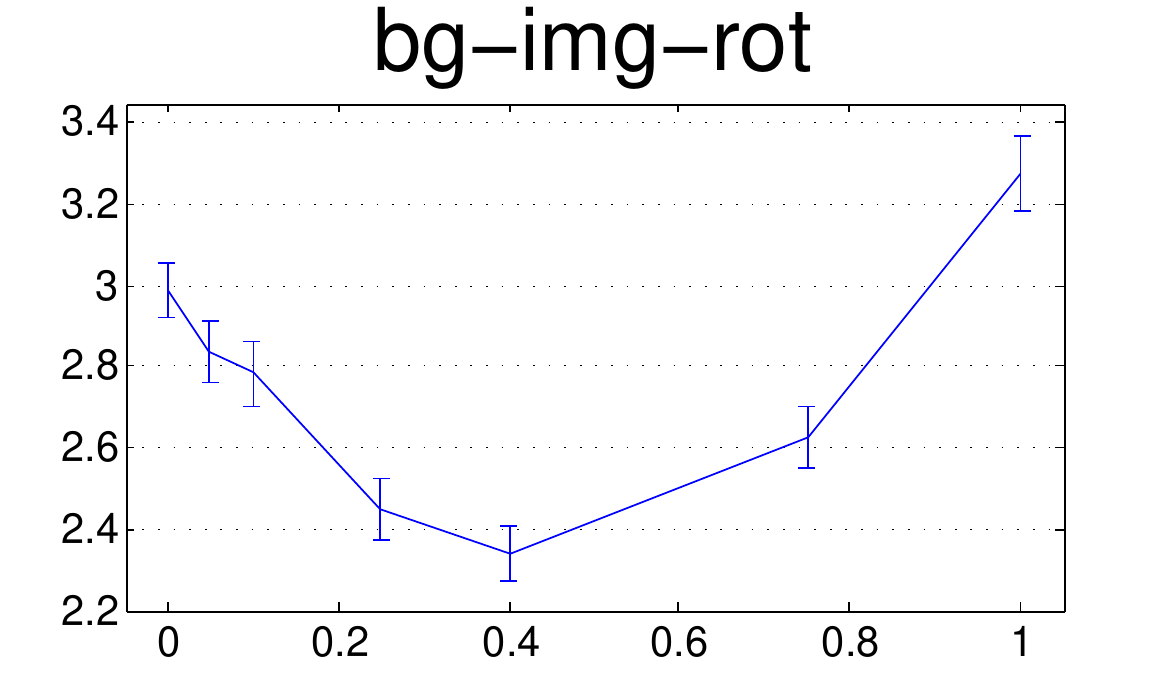}
\includegraphics[scale=0.36]{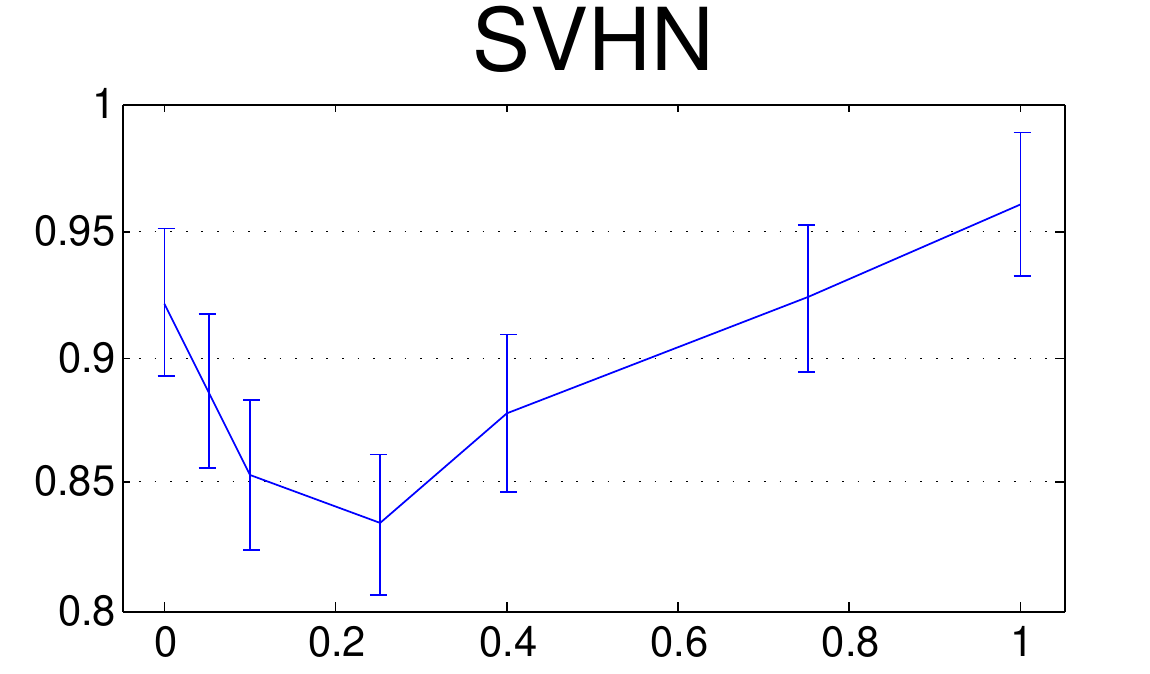}
\includegraphics[scale=0.365]{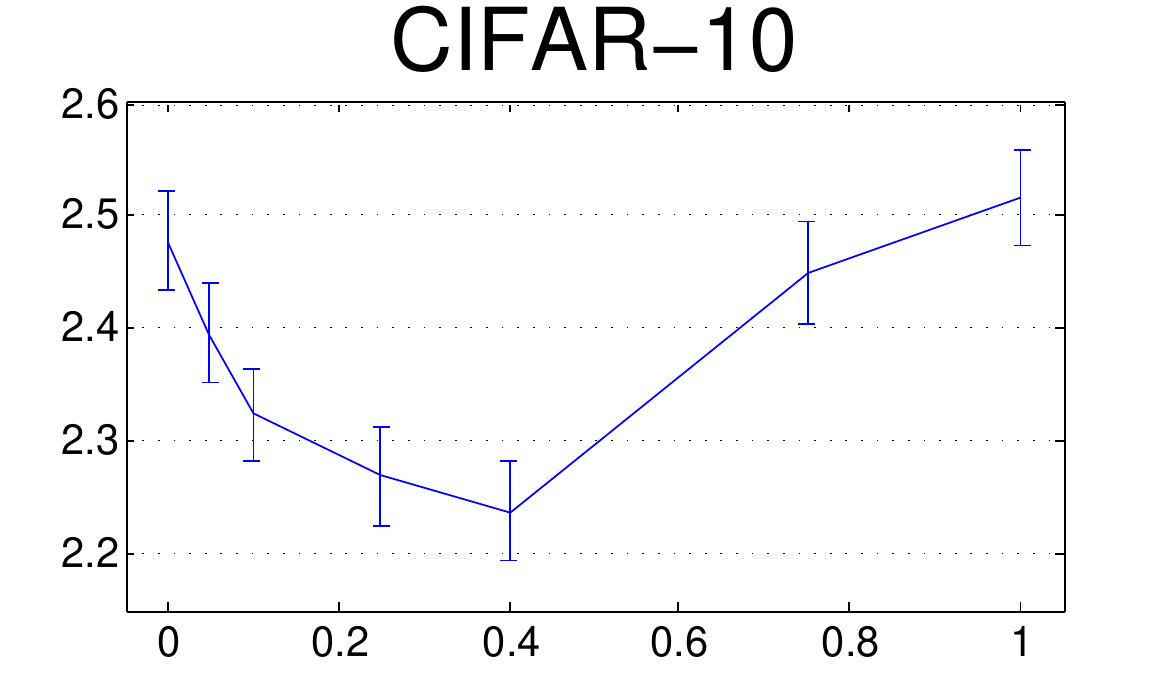}
\includegraphics[scale=0.36]{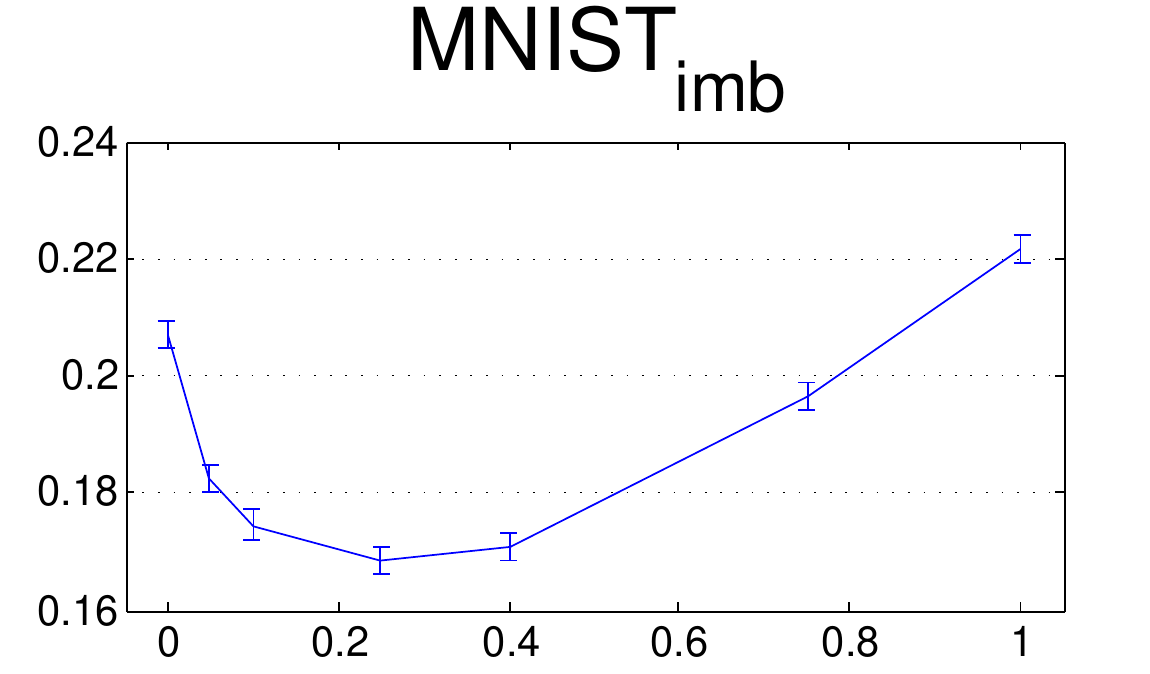}
\includegraphics[scale=0.36]{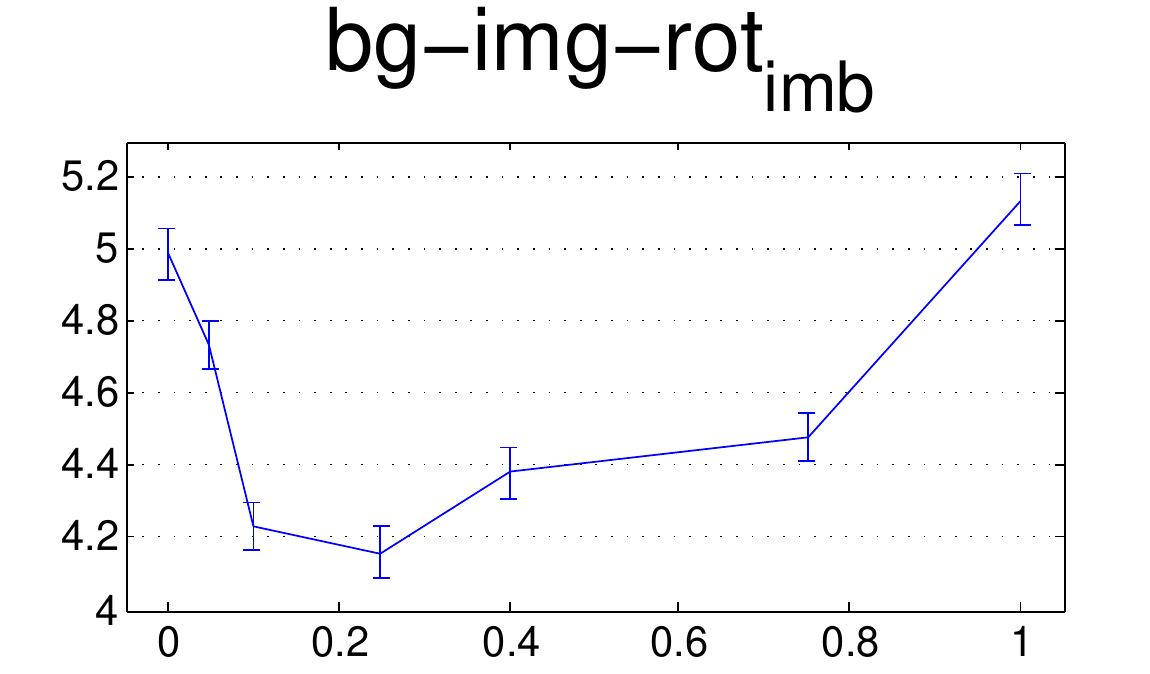}
\includegraphics[scale=0.36]{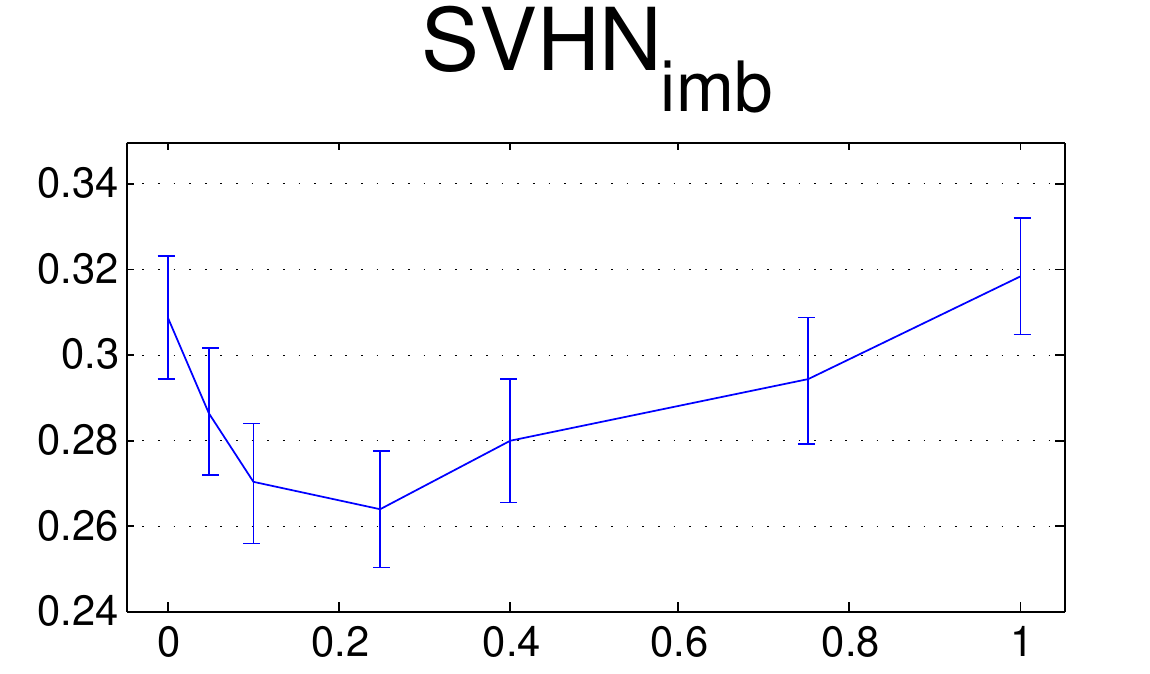}
\includegraphics[scale=0.365]{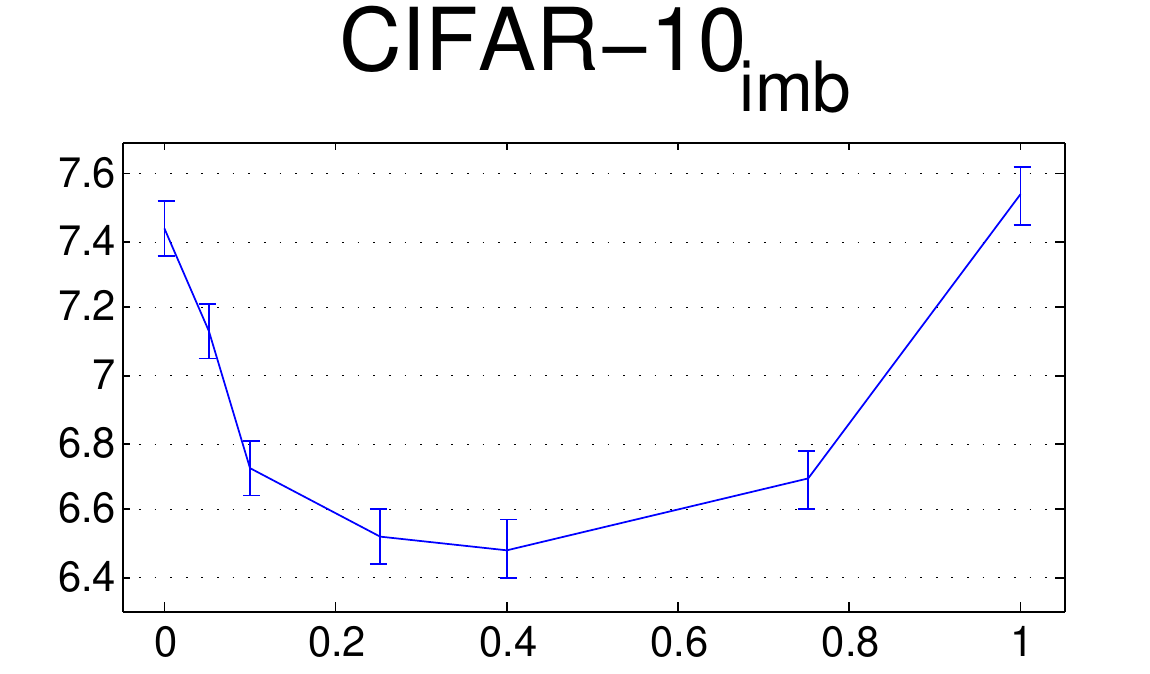}
\caption{Relation between $\beta$ and test cost (note that $\mathrm{SDAE}_{\mathrm{SOSR}}$ is the data point with $\beta=0$).}
\label{fg:beta_plot}
\end{figure}

\myparagraph{Is the mixture loss necessary?} To have more insights on $\mathrm{CAE}$, we also conducted experiments to evaluate the performance of $\mathrm{CSDNN}$ for $\beta\in[0, 1]$. When $\beta=0$, CSDNN degenerates to $\mathrm{SDAE}_{\mathrm{SOSR}}$; when $\beta=1$, each CAE of CSDNN performs fully cost-aware pre-training to fit the cost vectors. The results are displayed in Figure~\ref{fg:beta_plot}, showing a roughly U-shaped curve. The curve implies that some $\beta \in [0, 1]$ that best balances the tradeoff between denoising and cost-awareness can be helpful. The results validate the usefulness of the proposed mixture loss~(\ref{eq:cae_loss}) for pre-training.

\section{Conclusion}
\label{sec:conclusion}

We proposed a novel deep learning algorithm CSDNN for multiclass cost-sensitive classification with deep learning. Existing baselines and other alternatives within the $\mathrm{blind}$-series, the $\mathrm{Bayes}$-series and the $\mathrm{SOSR}$-series were extensively compared with CSDNN carefully to validate the importance of each component of CSDNN. The experimental results demonstrate that incorporating the cost information into both the pre-training and the training stages leads to promising performance of CSDNN, outperforming those baselines and alternatives. One key component of CSDNN, namely the $\mathrm{SOSR}$ loss for cost-sensitivity in the training stage, is shown to be helpful in improving the performance of CNN. The results justify the importance of the proposed $\mathrm{SOSR}$ loss for training and the CAE approach for pre-training.

\section{Acknowledgement}
We thank the anonymous reviewers for valuable suggestions. This material is based upon work supported by the Air Force Office of Scientific Research, Asian Office of Aerospace Research and Development (AOARD) under award number FA2386-15-1-4012, and by the Ministry of Science and Technology of Taiwan under number MOST 103-2221-E-002-149-MY3.

\bibliographystyle{named}
\bibliography{ijcai16}

\end{document}